\def\year{2021}\relax
\documentclass[letterpaper]{article} 
\usepackage{aaai21}  
\usepackage{times}  
\usepackage{helvet} 
\usepackage{courier}  
\usepackage[hyphens]{url}  
\usepackage{graphicx} 
\usepackage{natbib}  
\usepackage{caption} 
\usepackage[TABBOTCAP]{subfigure}
\usepackage[usenames, dvipsnames]{xcolor}
\usepackage{siunitx}        
\usepackage{bm}
\usepackage{tabularx}
\usepackage{makecell}
\usepackage{amsmath}
\usepackage{longtable}
\usepackage{multirow}
\usepackage{amsfonts}       
\usepackage{booktabs}       
\usepackage[shortlabels]{enumitem}  

\title{Parameterizing Branch-and-Bound Search Trees to Learn Branching Policies}

\author{
   Giulia Zarpellon,\textsuperscript{\rm 1}\thanks{Corresponding author. Code, data and supplementary materials can be found at: https://github.com/ds4dm/branch-search-trees}
   Jason Jo,\textsuperscript{\rm 2, 3}
   Andrea Lodi,\textsuperscript{\rm 1}
   Yoshua Bengio\textsuperscript{\rm 2, 3}\\
}
\affiliations{
    \textsuperscript{\rm 1}Polytechnique Montréal \\
    \textsuperscript{\rm 2}Mila \\
    \textsuperscript{\rm 3}Université de Montréal  

    giulia.zarpellon@polymtl.ca, jason.jo.research@gmail.com, andrea.lodi@polymtl.ca, yoshua.bengio@mila.quebec
}

\begin{document}
\maketitle
\begin{abstract}
  Branch and Bound (B\&B) is the exact tree search method typically used to solve Mixed-Integer Linear Programming problems (MILPs). Learning branching policies for MILP has become an active research area, with most works proposing to imitate the strong branching rule and specialize it to distinct classes of problems. We aim instead at learning a policy that generalizes across heterogeneous MILPs: our main hypothesis is that parameterizing the state of the B\&B search tree can aid this type of generalization. We propose a novel imitation learning framework, and introduce new input features and architectures to represent branching. Experiments on MILP benchmark instances clearly show the advantages of incorporating an explicit parameterization of the state of the search tree to modulate the branching decisions, in terms of both higher accuracy and smaller B\&B trees. The resulting policies significantly outperform the current state-of-the-art method for ``learning to branch'' by effectively allowing generalization to generic unseen instances.
\end{abstract}

\section{Introduction}
\label{sec:intro}

Many problems arising from transportation, healthcare, energy and logistics can be formulated as Mixed-Integer Linear Programming (MILP) problems, i.e., optimization problems in which some decision variables represent discrete or indivisible choices. 
A MILP is written as
\begin{equation}
    \label{eq:mip}
    \min_x\:\{ c^T x:  A x \ge b, x \ge 0,~x_i \in \mathbb{Z}~\forall i \in \mathcal{I}\},
\end{equation}
where $A\in\mathbb{R}^{m\times n}$, $b\in\mathbb{R}^m$, $c, x\in\mathbb{R}^n$ and $\mathcal{I}\subseteq\{1,\dots,n\}$ is the set of indices of variables that are required to be integral, while the other ones can be real-valued. 
Note that one can consider a MILP as defined by $(c, A, b, \mathcal{I})$; we do not assume any special combinatorial structure on the parameters $c, A, b$. 
While MILPs are in general $\mathcal{NP}$-hard, MILP solvers underwent dramatic improvements over the last decades \cite{lodi2010mixed,Achterberg2013} and now achieve high-performance on a wide range of problems. 
The fundamental component of any modern MILP solver is Branch and Bound (B\&B) \cite{LandDoig1960}, an \emph{exact} tree search method. 
Following a divide-and-conquer approach, B\&B partitions the search space by branching on variables' values and smartly uses bounds from problem relaxations to prune unpromising regions from the tree. 
The B\&B algorithm actually relies on expertly-crafted \emph{heuristic} rules for its two most fundamental decisions: \emph{branching variable selection} (BVS) and \emph{node selection}. 
In particular, BVS is a crucial factor for B\&B's success \cite{Achterberg2013}, and will be the main focus of the present article.

Understanding why B\&B works has been called ``one of the mysteries of computational complexity theory'' \cite{WhyDoesBBWork}, and there currently is no mathematical theory of branching; to the best of our knowledge, the only attempt in formalizing BVS is the recent work of \citeauthor{le2015abstract}. 
One central reason why B\&B is difficult to formalize resides in its inherent exponential nature: millions of BVS decisions could be needed to solve a MILP, and a single bad one could result in a doubled tree size and no improvement in the search. 
Such a complex and data-rich setting, paired with a lack of formal understanding, makes B\&B an appealing ground for machine learning (ML) techniques, which have lately been thriving in discrete optimization \cite{BLP18}. 
In particular, there has been substantial effort towards ``learning to branch'', i.e., in using ML methods to learn BVS policies \cite{Lodi2017}. 
Up to now, most works in this area focused on learning branching policies by supervision or imitation of \emph{strong branching} (SB), a valid but expensive heuristic scheme (see Sections~\ref{sec:background} and \ref{sec:related}).
The latest and state-of-the-art contribution to ``learning to branch'' \cite{NIPS2019_9690} frames BVS as a classification problem on SB expert decisions, and employs a graph-convolutional neural network (GCNN) to represent MILPs via their variable-constraint structure. 
The resulting branching policies improve on the solver by specializing SB to different classes of synthetic problems, and the attained generalization ability is to similar MILP instances (within the same class), possibly larger in formulation size.

The present work seeks a different type of generalization for a branching policy, namely a systematic generalization across \emph{heterogeneous} MILPs, i.e., across problems not belonging to the same combinatorial class, without any restriction on the formulation's structure and size. 
To achieve this goal, \emph{we parameterize BVS in terms of B\&B search trees}. 
On the one hand, information about the state of the B\&B tree -- abundant yet mostly unexploited by MILP solvers -- was already shown to be useful to learn resolution patterns shared across general MILPs \cite{MILPOutcome}. 
On the other hand, the state of the search tree ought to have a central role in BVS -- which ultimately decides how the tree is expanded and hence how the search itself proceeds. 
In practice, B\&B continually interacts with other algorithmic components of the solver to effectively search the decision tree, and some algorithmic decisions may be triggered depending on which phase the optimization is in \cite{BertholdHendelKoch2017}. 
In a highly integrated framework, a branching variable should thus be selected among the candidates based on its role in the search and its various components. 
Indeed, state-of-the-art heuristic branching schemes employ properties of the tree to make BVS decisions, and the B\&B method equipped with such branching rules has proven to be successful across widely heterogeneous instances. 

Motivated by these considerations, our main hypothesis is that MILPs share a higher order structure in the space of B\&B search trees, and parameterized BVS policies should learn in this representational space. 
We setup a novel learning framework to investigate this idea. First of all, there is no natural input representation of this underlying space. 
Our first contribution is to craft input features of the variables that are candidates for branching: we aim at representing their roles in the search and its dynamic evolution. 
The dimensionality of such descriptions naturally changes with the number of candidates at every BVS step. 
The deep neural network (DNN) architecture that we propose learns a baseline branching policy (NoTree) from the candidate variables' representations and effectively deals with varying input dimensions. 
Taking this idea further, we suggest that an explicit representation of the \emph{state of the search tree} should condition the branching criteria, in order for it to flexibly adapt to the tree evolution. 
We contribute such tree-state parameterization, and incorporate it to the baseline architecture to provide context over the candidate variables at each given branching step. 
In the resulting policy (TreeGate) the tree state acts as a control mechanism to drive a top-down modulation (specifically, feature gating) of the highly mutable space of candidate variables representations. 
By training in our hand-crafted input space, the signal-to-noise ratio of the high-level branching structure shared amongst general MILPs is effectively increased, enabling our TreeGate policy to rapidly infer these latent factors and dynamically compose them via top-down modulation. 
In this sense, \emph{we learn branching from parameterizations of B\&B search trees that are shared among general MILPs}.
To the best of our knowledge, the present work is the first attempt in the “learning to branch” literature to represent B\&B search trees for branching, and to establish such a broad generalization paradigm covering many classes of MILPs. 

We perform imitation learning (IL) experiments on a curated dataset of heterogeneous instances from standard MILP benchmarks. 
We employ as expert rule \texttt{relpscost}, the default branching scheme of the optimization solver SCIP \cite{GleixnerEtal2018ZR}, to which our framework is integrated. 
Machine learning experimental results clearly show the advantage of the policy employing the tree state (TreeGate) over the baseline one (NoTree), the former achieving a 19\% improvement in test accuracy. 
When plugged in the solver, both learned policies compare well with state-of-the-art branching rules. 
The evaluation of the trained policies in the solver also supports our idea that representing B\&B search trees enables learning to branch across generic MILP instances: over test instances, the best TreeGate policy explores on average trees with 27\% less nodes than the best NoTree one. 
In contrast, the GCNN framework of \citeauthor{NIPS2019_9690} that we use as benchmark does not appear to be able to attain such broad generalization goal: often the GCNN models fail to solve heterogeneous test instances, exploring search trees that are considerably bigger than those we obtain. 
The comparison thus remarks the advantage of our fundamentally new architectural paradigm -- of representing candidates’ role in the search and using a tree-context to modulate BVS -- which without training in a class-specific manner nor focusing on constraints structure effectively allows learning across generic MILPs.
 
\section{Background}
\label{sec:background}

Simply put, the B\&B algorithm iteratively partitions the solution space of a MILP (\ref{eq:mip}) into sub-problems, which are mapped to nodes of a binary decision tree. 
At each node, integrality requirements for variables in $\mathcal{I}$ are dropped, and a linear programming (LP) (continuous) relaxation of the problem is solved to provide a valid lower bound to the optimal value of (\ref{eq:mip}). 
When the solution $x^*$ of a node LP relaxation violates the integrality of some variables in $\mathcal{I}$, that node is further partitioned into two children by \emph{branching on a fractional variable}. 
Formally, $\mathcal{C}=\{i\in\mathcal{I}: x_i^*\notin\mathbb{Z}\}$ defines the index set of \emph{candidate variables} for branching at that node. 
The BVS problem consists in selecting a variable $j\in\mathcal{C}$ in order to \emph{branch} on it, i.e., create child nodes according to the split
\begin{equation}\label{eq:split}
    x_j\le\lfloor x_j^* \rfloor \:\vee\: x_j\ge\lceil x_j^* \rceil.
\end{equation}
Child nodes inherit a lower bound estimate from their parent, while (\ref{eq:split}) ensures $x^*$ is removed from their solution spaces. 
After extending the tree, the algorithm moves on to select a new \emph{open node}, i.e., a leaf yet to be explored (node selection): a new relaxation is solved, and new branchings happen. When $x^*$ satisfies integrality requirements, then it is actually feasible for (\ref{eq:mip}), and its value provides a valid upper bound to the optimal one. 
Maintaining global upper and lower bounds allows one to prune large portions of the search space. 
During the search, \emph{final leaf nodes} are created in three possible ways: by integrality, when the relaxed solution is feasible for (\ref{eq:mip}); by infeasibility of the sub-problem; by bounds, when the comparison of the node's lower bound to the global upper one proves that its sub-tree is not worth exploring. 
An optimality certificate is reached when the global bounds converge. 
See \cite{Wolsey1998,lodi2010mixed} for details on B\&B and its combination with other components of a MILP solver.

\paragraph{Branching rules}{
Usually, candidates are evaluated with respect to some scoring function, and $j$ is chosen for branching as the (or a) score-maximizing variable \cite{AKM05}. 
The most used criterion in BVS measures variables depending on the improvement of the lower bound in their (prospective) child nodes. 
The \emph{strong branching} (SB) rule \cite{ApplegateSB} explicitly computes bound gains for $\mathcal{C}$. 
The procedure is expensive, but experimentally realizes trees with the least number of nodes. 
Instead, \emph{pseudo-cost} (PC) \cite{BGGH71} maintains a history of variables' branchings, averaging past improvements to get a proxy for the expected gain. 
Fast in evaluation, PC can behave badly due to uninitialization, so combinations of SB with PC have been developed. 
In \emph{reliability branching}, SB is performed until PC scores for a variable are deemed reliable proxies of bound improvements. 
In \emph{hybrid branching} \cite{achterberg_hybrid_2009}, PC scores are combined with other ones measuring the variables' role on inference and conflict clauses. 
Many other scoring criteria have been proposed, and some of them are surveyed in \citeauthor{Lodi2017} from a ML perspective.

State-of-the-art branching rules can in fact be interpreted as mechanisms to score variables based on their effectiveness in different search components. 
While hybrid branching explicitly combines five scores reflecting variables' behaviors in different search tasks, the evaluation performed by SB and PC can also be seen as a measure of how effective a variable is -- in the single task of improving the bound from one parent node to its children. 
Besides, one can assume that the importance of different search functionalities should change dynamically during the tree exploration.
\footnote{Indeed, a ``dynamic factor'' adjusts weights in the default branching scheme of SCIP \cite{relpscost}.}
In this sense, our approach aims at learning a branching rule that takes into account variables' roles in the search and the tree evolution itself to perform a more flexible BVS, adapted to the search stages.
}

\section{Parameterizing B\&B Search Trees}
\label{sec:representations}

The central idea of our framework is to learn BVS by means of parameterizing the underlying space of B\&B search trees. 
We believe this space can represent the complexity and the dynamism of branching in a way that is shared across heterogeneous problems. 
However, there are no natural parameterization of BVS or B\&B search trees. 
To this end, our contribution is two-fold: 1) we propose hand-crafted input features to describe candidate variables in terms of their roles in the B\&B process, and explicitly encode a ``tree state'' to provide a richer context to variable selection; 2) we design novel DNN architectures to integrate these inputs and learn BVS policies.

\subsection{Hand-crafted Input Features}
\label{subsec:features}

\begin{figure*}[!t]
    \centering
    \subfigure[\label{fig:tsne-eil33-2}]{\includegraphics[width=.3\linewidth]{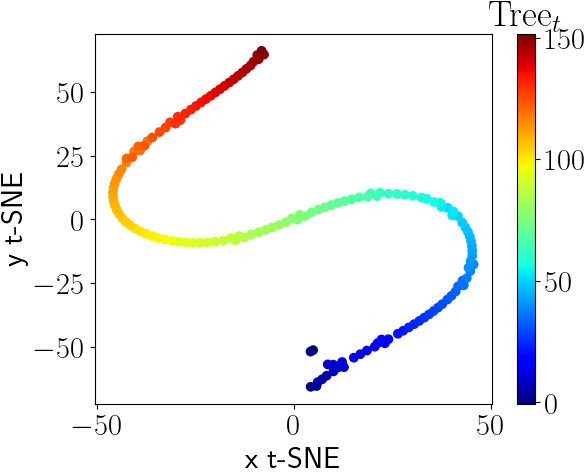}}\quad
    \subfigure[\label{fig:tsne-seymour1}]{\includegraphics[width=.3\linewidth]{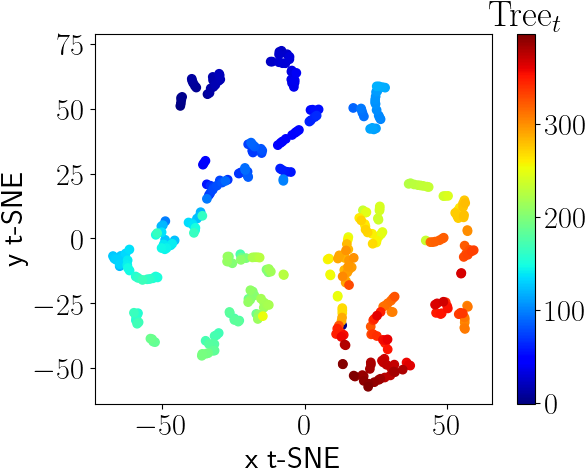}}\quad
    \subfigure[\label{fig:cands-hist}]{\includegraphics[width=.33\linewidth]{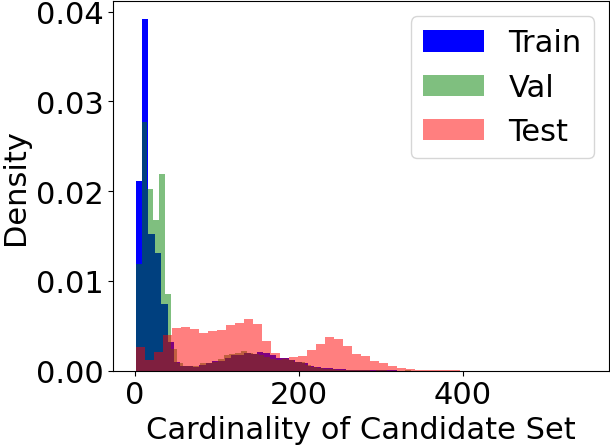}}
    \caption{Evolution of $\mathit{Tree}_t$ throughout B\&B as synthesized by t-SNE plots (perplexity=5), for instances (a) eil33-2 and (b) seymour1. (c) Histogram of $|\mathcal{C}_t|$ in train, validation and test data.}
    \label{fig:repr-plots}
\end{figure*}

At each branching step $t$, we represent the set of variables that are candidates for branching by an input matrix $C_t\in\mathbb{R}^{25\times |\mathcal{C}_t|}$. 
To capture the multiple roles of a variable throughout the search, we describe each candidate $x_j, j\in\mathcal{C}_t$ in terms of its bounds and solution value in the current sub-problem. 
We also feature statistics of a variable's participation in various search components and in past branchings. 
In particular, the scores that are used in the SCIP default hybrid-branching formula are part of $C_t$.

Additionally, we create a separate parameterization $\mathit{Tree}_t\in\mathbb{R}^{61}$ to describe the state of the search tree. 
We record information of the current node in terms of depth and bound quality. 
We also consider the growth rate and the composition of the tree, the evolution of global bounds, aggregated variables' scores, statistics on feasible solutions and on bound estimates and depths of open nodes.

All features are designed to capture the dynamics of the B\&B process linked to BVS decisions, and are efficiently gathered through a customized version of PySCIPOpt \cite{pyscipopt}. 
Note that $\{C_t, \mathit{Tree}_t\}$ are defined in a way that is not explicitly dependent on the parameters of each instance $(c, A, b, \mathcal{I})$. 
Even though $C_t$ naturally changes its dimensionality at each BVS step $t$ depending on the highly variable $\mathcal{C}_t$, the fixed lengths of the vectors enable training among branching sets of different sizes (see \ref{subsec:architecture}). 
The representations evolve with the search: t-SNE plots \cite{t-SNE} in Figures~\ref{fig:tsne-eil33-2} and \ref{fig:tsne-seymour1} synthesize the evolution of $\mathit{Tree}_t$ throughout the B\&B search, for two different MILP instances. 
The pictures clearly show the high heterogeneity of the branching data across different search stages. 
A detailed description of the hand-crafted input features is reported in the supplementary material (SM). 

\subsection{Architectures to Model Branching}
\label{subsec:architecture}

\begin{figure}[!t]
    \centering
    \includegraphics[width=.95\linewidth]{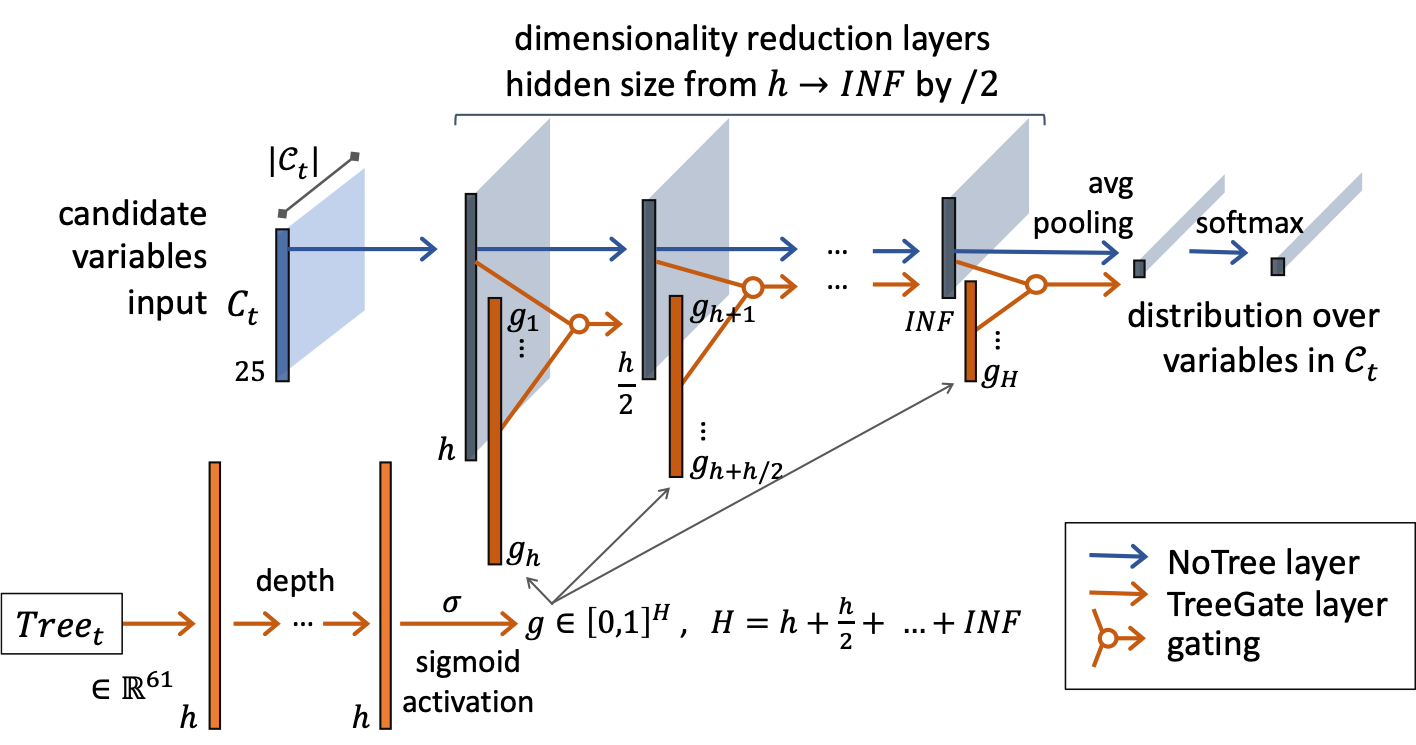} 
    \caption{Candidate variables input $C_t$ is processed by NoTree layers (in blue) to select a variable for branching. For the TreeGate model, the $\mathit{Tree}_t$ input is first embedded and then utilized in gating layers (in orange) on the candidates' representations.}
    \label{fig:treegate}
\end{figure}

We use parameterizations $C_t$ as inputs for a baseline DNN architecture (NoTree). 
Referring to Figure~\ref{fig:treegate}, the 25-feature input of a candidate variable is first embedded into a representation with hidden size $h$; subsequently, multiple layers reduce the dimensionality from $h$ to an infimum $\mathit{INF}$ by halving it at each step. 
The vector of length $\mathit{INF}$ is then compressed by global average pooling into a single scalar. 
The $|\mathcal{C}_t|$ dimension of $C_t$ is conceived (and implemented) as a ``batch dimension'': this makes it possible to handle branching sets of varying sizes, still allowing the parameters of the nets to be shared across problems. 
Ultimately, a $\mathit{softmax}$ layer yields a probability distribution over the candidate set $\mathcal{C}_t$, according to which a variable is selected for branching.

We incorporate the tree-state input to the baseline architecture to provide a search-based context over the mutable branching sets. 
Practically, $\mathit{Tree}_t$ is embedded in a series of subsequent layers with hidden size $h$. 
The output of a final sigmoid activation is $g\in[0,1]^{H}$, where $H=h+h/2+\dots+\mathit{INF}$ denotes the total number of units of the NoTree layers. 
Separate chunks of $g$ are used to modulate by feature gating the representations of NoTree: $[g_1, \dots, g_h]$ controls features at the first embedding, $[g_{h+1}, \dots, g_{h+h/2}]$ acts at the second layer, \dots, and so on, until exhausting $[g_{H-\mathit{INF}}, \dots, g_H]$ with the last layer prior the average pooling. 
In other words, $g$ is used as a control mechanism on variables parameterization, gating their features via a learned tree-based signal. 
The resulting network (TreeGate) models the high-level idea that a branching scheme should \emph{adapt to the tree evolution}, with variables' selection criteria dynamically changing throughout the tree search.

\subsection{Imitation Learning}
\label{subsec:imitationlearning}

We train our BVS policies via imitation learning, specifically behavioral cloning \cite{BehavioralCloning}. 
Our expert branching scheme is SCIP default, \texttt{relpscost}, i.e., a reliability version of hybrid branching in which SB and PC scores are complemented with other ones reflecting the candidates' role in the search; \texttt{relpscost} is a more realistic expert (nobody uses SB in practice), and the most suited in our context, given the emphasis we put on the search tree. 
One of the main challenges of learning to imitate a complex BVS expert policy is that it is only possible to partially observe the true state of the solver. 
In our learning framework, we approximate the solver state with our parameterized B\&B search tree state $\mathrm{x}_t = \{C_t, \mathit{Tree}_t\}$ at branching step $t$. 
For each MILP instance, we roll-out SCIP to gather a collection of pairs $\mathcal{D} = (\mathrm{x}_t, \mathrm{y}_t)_{t=1}^N$ where $\mathrm{y}_t \in\mathcal{C}_t$ is the branching decision made by \texttt{relpscost} at branching step $t$. 
Our policies $\pi_{\theta}$ are trained to minimize the cross-entropy categorical loss function: 
\begin{equation}\label{eq:il-objective}
   \mathcal{L}(\theta) = -\frac{1}{N}\sum_{ (\mathrm{x}, \mathrm{y}) \in \mathcal{D}} \log \pi_{\theta}(\mathrm{y} \mid \mathrm{x}).
\end{equation}

\subsection{Systematic Generalization}
\label{subsec:systematicgen}

Our aim is to explore systematic generalization in learning to branch. 
We measure systematic generalization by evaluating how well trained policies perform on never-seen heterogeneous MILP instances. 
To begin, we remark that \texttt{relpscost} is a sophisticated ensemble of expertly-crafted heuristic rules that are dependent on high-level branching factors. 
Due to the systematic generalization failures currently plaguing even state-of-the-art DNNs \cite{LakeSystematic, CLEVRCOGENT, GoodfellowAdvExs}, we do not believe that current learning algorithms are capable of inferring these high-level branching factors from the raw MILP formulation data alone. 
Instead, by opting to train BVS policies in our hand-crafted input feature space: (1) the signal-to-noise ratio of the underlying branching factors is effectively increased, and (2) the likelihood of our models overfitting to superficial regularities is vastly decreased as this tree-oriented feature space is abstracted away from instance or class specific peculiarities of the MILP formulation. 
However, in order to achieve systematic generalization, inference of high-level latent factors must be paired with a corresponding composition mechanism. 
Here, the top-down neuro-modulatory prior of our TreeGate model represents a powerful mechanism for dynamically composing the tree state and the candidate variable states together. 
In summary, we hypothesize that BVS policies trained in our framework are able to better infer and compose latent higher order branching factors, which in turn enables flexible generalization across heterogeneous MILP instances. 

\section{Experiments}
\label{sec:experiments}

Our experiments are designed to carefully measure the inductive bias of BVS learning frameworks. 
We echo the sentiment of \citeauthor{MogrifierLSTM} that merely scaling up a dataset is insufficient to explore the issue of systematic generalization in deep learning models, and we instead choose to curate a controlled dataset with the following properties: 
\begin{itemize}
    \item[-] \emph{Train/test split consisting of heterogeneous MILP instances.} 
    This rules out the possibility of BVS policies learning superficial class-specific branching rules and instead forces them to infer higher order branching factors shared across MILPs.  
    
    \item[-] \emph{A restricted set of training instances.} 
    While limiting the sample complexity of the training set poses a significant learning challenge, the learning framework with the best inductive bias will be the one that best learns to infer and compose higher order branching factors from such a difficult training set.
\end{itemize}

While we train our BVS policies to imitate the SCIP default policy \texttt{relpscost}, our objective is not to outperform \texttt{relpscost}. 
Indeed, expert BVS policies like \texttt{relpscost} are tuned over \emph{thousands} of solvers' proprietary instances: comprehensively improving on them is a very hard task -- impossible to guarantee in our purely-research experimental setting -- and should thus not be the only yardstick to determine the validity (and practicality) of a learned policy. 
Instead, we focus on evaluating how efficiently BVS learning frameworks can learn to infer and compose higher order branching factors by measuring generalization from a controlled training set to heterogeneous MILP test instances.  

\paragraph{MILP dataset and solver setting}{
In general, randomly-generated \emph{generic} MILPs are too easy to be of interest; besides, public MILP libraries only contain few hundreds of instances, not all viable for our setting, and a careful dataset curation is thus needed. 
Comparisons of branching policies become clearer when the explored trees are manageable in size and the problems can be consistently solved to optimality. 
Thus we select 27 heterogeneous problems from real-world MILP benchmark libraries \cite{miplib3,MIPLIB2010,miplib2017,milplib}, focusing on instances whose tree exploration is on average relatively contained (in the tens/hundreds of thousands nodes, max.) and whose optimal value is known. 
We partition our selection into 19 train and 8 test problems, which are listed in Table~\ref{tab:milp_dataset} (see SM for more details). 

We use SCIP 6.0.1. 
Modifying the solver configuration is common practice in BVS literature \cite{LinderothSavelsbergh}, especially in a proof-of-concept setting in which our work is positioned. 
To reduce the effects of the other solver's components on BVS, we work with a configuration specifically designed to fairly compare the performance of branching rules \cite{gamrath_impact}. 
In particular, we disable all primal heuristics and for each problem we provide the known optimal solution value as cutoff. 
We also enforce a time-limit of 1h. 
Further details on the solver parameters and hardware settings are reported in the SM.
}

\begin{table}[t]
\centering
\subtable[\label{tab:milp_dataset}]{
\begin{small}
    \begin{tabular}{p{0.9\linewidth}}
    \toprule
        \textsc{Train}: air04, air05, dcmulti, eil33-2, istanbul-no-cutoff, l152lav, lseu, misc03, neos20, neos21, neos-476283, neos648910, pp08aCUTS, rmatr100-p10, rmatr100-p5, sp150x300d, stein27, swath1, vpm2\\
    \midrule
        \textsc{Test}: map18, mine-166-5, neos11, neos18, ns1830653, nu25-pr12, rail507, seymour1\\
    \bottomrule
    \end{tabular}
\end{small}
}\hfill
\subtable[\label{tab:split}]{
\begin{small}
    \begin{tabular}{lcl}
    \toprule
         & \textbf{Total} & \textbf{$(s, k)$ pairs}\\
    \midrule
        Train & 85,533 & $\{0, 1, 2, 3\}\times\{0, 1, 5, 10, 15\}$\\
        Valid. & 14,413 & $\{4\}\times\{0, 1, 5, 10, 15\}$\\
        Test & 28,307 & $\{0, 1, 2, 3, 4\}\times\{0\}$\\
    \bottomrule
    \end{tabular}
\end{small}
}
\caption{(a) List of MILP instances in train and test sets. (b) For train, validation and test set splits we report the total number of data-points and the seed-$k$ pairs $(s,k)$ from which they are obtained.}
\end{table}

\paragraph{Data collection and split}{
We collect IL training data from SCIP roll-outs, gathering inputs $\mathrm{x}_t = \{C_t, \mathit{Tree}_t\}$ and corresponding \texttt{relpscost} branching decisions (labels) $\mathrm{y}_t \in\mathcal{C}_t$. 
Given that each branching decision gives rise to a single data-point $(\mathrm{x}_t, \mathrm{y}_t)$, and that the search trees of the selected MILP instances are not extremely big, one needs to augment the data. 
We proceed in two ways.
\begin{enumerate}[align=left]
    \item[i.] We exploit MILPs \emph{performance variability} \cite{LT13}, and obtain perturbed searches of the same instance by setting solver's random seeds $s\in\{0,...,4\}$ to control variables' permutations.
    
    \item[ii.] We diversify B\&B explorations by running the branching scheme \texttt{random} for the first $k$ nodes, before switching to SCIP default rule and starting data collection. 
    The motivation behind this type of augmentation is to gather input states that are unlikely to be observed by an expert rule \cite{LearningToSearch}. 
    We select $k\in\{0, 1, 5, 10, 15\}$, where $k=0$ corresponds to a run without random branching. We apply this type of augmentation to \emph{train instances only}.
\end{enumerate}
One can quantify MILP variability by computing the coefficient of variation of the performance measurements \cite{MIPLIB2010}; 
we report such scores and measure the effect of $k$ initial random branchings in the SM. 
Overall, both (i) and (ii) appear effective to diversify our dataset. 
The final composition of train, validation and test sets is summarized in Table~\ref{tab:split}. 
Train and validation data come from the same instances; the test set contains samples from separate MILPs, using only type (i) augmentations.

\begin{table}[!t]
    \centering
    \begin{footnotesize}
    \begin{tabular}{llccc}
        \toprule
           Policy & $h$ / $d$ / $\mathit{LR}$ & Test acc (@5) & Val acc (@5) \\
        \toprule
            NT & 128 / -- / 0.001 & 64.02 (88.51) & 77.69 (95.88) \\
            TG & 64 / 5 / 0.01 & 83.70 (95.83) & 84.33 (96.60)  \\
        \midrule
            \textsc{gcnn} &  -- / -- / --  & 15.28 (44.16) & 19.28 (38.44)  \\
        \bottomrule
    \end{tabular}
    \end{footnotesize}
    \caption{Selected NoTree (NT) and TreeGate (TG) models with corresponding hyper-parameters, and test and validation accuracy. For \textsc{gcnn} we report \emph{average} scores across 5 seeds; validation means use the best scores observed during training.}
    \label{tab:best_policies_gcnn}
\end{table}

An important measure to analyze the dataset is given by the size of the candidate sets (i.e., the varying dimensionality of the $C_t$ inputs) contained in each split. 
Figure~\ref{fig:cands-hist} shows histograms for $|\mathcal{C}_t|$ in each subset. 
While in train and validation the candidate set sizes are mostly concentrated in the $[0,50]$ range, the test set has a very different distribution of $|\mathcal{C}_t|$, and in particular one with a longer tail (over 300). 
In this sense, the test instances present never-seen branching data gathered from heterogeneous MILPs, and we test the generalization of our policies to \emph{entirely unknown and larger branching sets}.
}

\paragraph{IL optimization and GCNN benchmark}{
We train both IL policies using ADAM \cite{ADAM} with default $\beta_1=0.9, \beta_2=0.999$, and weight decay \num{1e-5}. 
Our hyper-parameter search spans: learning rate $\mathit{LR}\in\{0.01, 0.001, 0.0001\}$, hidden size $h\in\{32, 64, 128, 256\}$, and depth $d\in\{2, 3, 5\}$. 
The factor by which units of NoTree are reduced is 2, and we fix $\mathit{INF}=8$. 
We use PyTorch \cite{PyTorch} to train the models for 40 epochs, reducing $\mathit{LR}$ by a factor of 10 at epochs 20 and 30. 
To benchmark our results, we also train the GCNN framework of \citeauthor{NIPS2019_9690} on our MILP dataset. 
Data collection and experiments are carried out as in \citeauthor{NIPS2019_9690}, with full SB as expert, but we fix the solver setting as discussed above.
}

\subsection{Results}
\label{sec:results}

In our context, standard IL metrics are informative yet incomplete measures of performance for a learned BVS model, and one also cares about assessing the policies' behaviors when plugged in the solver environment. 
This is why in order to determine the best NoTree and TreeGate policies we take into account both types of evaluations. 
We first select few policies based on their test accuracy score; next, we specify them as custom branching rules in SCIP and perform full roll-outs on the entire MILP dataset, over five random seeds (i.e., 135 evaluations each). 
To summarize the policies' performance in the solver, we compute the shifted geometric mean (with a shift of 100) of the total number of nodes, over the 135 B\&B executions (\textsc{All}), and restricted to \textsc{Train} and \textsc{Test} instances. 

Both types of metrics are extensively reported in the SM, together with the policies' hyper-parameters. 
Incorporating an explicit parameterization of the state of the search tree to modulate BVS clearly aids generalization: the advantage of TreeGate over NoTree is evident in all metrics, and across multiple trained policies. 
What we observe is that best test accuracy does not necessarily translate into best solver performance. 
We select as best policies those yielding the best nodes average over the entire dataset (Table~\ref{tab:best_policies_gcnn}). 
In the case of TreeGate, the best model corresponds to that realizing the best top-1 test accuracy (83.70\%), and brings a 19\% (resp. 7\%) improvement over the NoTree policy, in top-1 (resp. top-5) test accuracy. 
The \textsc{gcnn} models (trained, tested and evaluated over 5 seeds, as in \citeauthor{NIPS2019_9690}) struggle to fit data from heterogeneous instances: their average top-1 (resp. top-5) test accuracy is only 15.28\% (resp. 44.16\%), and across \textsc{All} instances they explore around three times the number of nodes needed by our policies. 
Note that \textsc{gcnn} is a memory intensive model, and we had to drastically reduce the batch size parameter to avoid memory issues when using our instances. 
Learning curves and further details on training dynamics and test results can be found in the SM. 

\begin{table*}[!t]
    \centering
    \begin{footnotesize}
        \begin{tabular}{lrrrrrr}
        \toprule
            Instance &    NoTree &  TreeGate &  \textsc{gcnn} &     \texttt{random} &     \texttt{pscost} &  \texttt{relpscost} (fair) \\
        \midrule
            \textsc{All} &    1241.79 &   1056.79 &  *3660.32 &    *6580.79 &    *1471.61 &     286.15 (719.20) \\
            \textsc{Train} &     834.40 &    759.94 &   *1391.41 &    *2516.04 &     884.37 &     182.27 (558.34) \\
            \textsc{Test} &    3068.96 &   2239.47 &  *33713.63 &   *61828.29 &    *4674.34 &     712.77 (1276.76) \\
        \midrule
              map18 &       457.89 &    575.92 &  *3907.64 &   11655.33 &    1025.74 &     270.25 (441.18) \\
         mine-166-5 &      3438.44 &   4996.48 &  *233142.25 &  *389437.62 &    4190.41 &     175.10 (600.22) \\
             neos11 &      3326.32 &   3223.46 &  1642.07 &   29949.69 &    4728.49 &    2618.27 (5468.05) \\
             neos18 &     15611.63 &  10373.80 &  40794.74 &  228715.62 &  *133437.40 &    2439.29 (5774.36) \\
          ns1830653 &      6422.37 &   5812.03 &  *22931.45 &  288489.30 &   12307.90 &    3489.07 (4311.84) \\
          nu25-pr12 &       357.00 &     86.80 &  *45982.34 &    1658.41 &     342.47 &      21.39 (105.61) \\
            rail507 &      9623.05 &   3779.05 &  *75663.48 &   *80575.84 &    4259.98 &     543.39 (859.37) \\
           seymour1 &      3202.20 &   1646.82 &  *319046.04 &  *167725.65 &    3521.47 &     866.32 (1096.67) \\
        \bottomrule
    \end{tabular}
\end{footnotesize}
\caption{Total number of nodes explored by learned and SCIP policies for test instances and aggregated over sets, in shifted geometric means over 5 runs on seeds $\{0, \dots, 4\}$. We mark with * the cases in which time-limits were hit. For \texttt{relpscost}, we also compute the \emph{fair} number of nodes.}
    \label{tab:nnodes_testeval}
\end{table*}

In solver evaluations, NoTree and TreeGate are also compared to SCIP default branching scheme \texttt{relpscost}, PC branching \texttt{pscost} and a \texttt{random} one. 
For \texttt{relpscost} we also compute the \emph{fair} number of nodes \cite{gamrath_impact}, which accounts for those nodes that are processed as side-effects of SB-like explorations, specifically looking at domain reduction and cutoffs counts. 
In other words, the fair number distinguishes tree-size reductions due to better branching from those obtained by SB side-effects. 
For rules that do not involve any SB, the fair number and the usual nodes' count coincide. 
The selected solver parametric setting (the same used for data collection and \textsc{gcnn} benchmark) allows a meaningful computation of the fair number of nodes, and a honest comparison of branching schemes. 

Both NoTree and TreeGate policies are able to solve all instances within the 1h time-limit, like \texttt{relpscost}. 
In contrast, \textsc{gcnn} hits the limit on 7 instances (24 times in total), while \texttt{random} does so on 4 instances (17 times in total) and \texttt{pscost} on one instance only (neos18), a single time. 
Table~\ref{tab:nnodes_testeval} reports the nodes' means for every test instance over five runs (see SM for complete instance-specific results), as well as measures aggregated over train and test sets, and the entire dataset. 
In aggregation, TreeGate is always better than NoTree, the former exploring on average trees with 14.9\% less nodes. 
This gap becomes more pronounced when measured over test instances only (27\%), indicating the advantage of TreeGate over NoTree when exploring unseen data. 
Results are less clear-cut from an instance-wise perspective, with neither policy emerging as an absolute winner, though the reductions in tree sizes achieved by TreeGate are overall more pronounced. 
While the multiple time-limits of \textsc{gcnn} hinder a proper comparison in terms of explored nodes, results clearly indicate that the difficulties of \textsc{gcnn} exacerbate over never-seen, heterogeneous test instances.

Our policies also compare well to other branching rules: both NoTree and TreeGate are substantially better than \texttt{random} across all instances, and always better than \texttt{pscost} in aggregated measures. 
Only on one training instance both policies are much worse than \texttt{pscost} (neos-476283); in the test set, \textsc{gcnn} appears competitive with our models only on neos11.
As expected, \texttt{relpscost} still realizes the smallest trees, but on 11 (out of 27) instances at least one among NoTree and TreeGate explores less nodes than the \texttt{relpscost} fair number. 
In general, our policies realize tree sizes comparable to the SCIP ones, when SB side effects are taken into account.

\section{Related Work}
\label{sec:related}

Among the first attempts in ``learning to branch'', \citeauthor{Alvarez2017} perform regression to learn proxies of SB scores. 
Instead, \citeauthor{Khalil2016} propose to learn the ranking associated with such scores, and train instance-specific models (that are not end-to-end policies) via $\text{SVM}^{\text{rank}}$. 
Also \citeauthor{hansknecht2018cuts} treat BVS as a ranking problem, and specialize their models to the combinatorial class of time-dependent traveling salesman problems. 
More recently, the work of \citeauthor{BalcanICML2018} learns mixtures of existing branching schemes for different classes of synthetic problems, focusing on sample complexity guarantees. 
In \citeauthor{liberto_dash_2016}, a portfolio approach to BVS is explored. 
Similarly to us, \citeauthor{NIPS2019_9690} frames BVS as classification of SB-expert branching decisions and employs a GCNN model to learn branching. 
Proposed features in \citeauthor{NIPS2019_9690} (as in \citeauthor{Alvarez2017,Khalil2016}) focus on static, parameters-dependent properties of MILPs and node LP relaxations, whereas our representations aim at capturing the temporality and dynamism of BVS. 
Although their resulting policies are specializations of SB that appear to effectively capture structural characteristics of some classes of combinatorial optimization problems, and are able to generalize to larger formulations from the same distribution, we showed how such policies fail to attain a broader generalization paradigm.

Still concerning the B\&B framework, \citeauthor{LearningToSearch} employ IL to learn a heuristic class-specific node selection policy; \citeauthor{song2018learning} propose instead a retrospective approach on IL. 
A reinforcement learning (RL) approach for node selection can be found in \citeauthor{sabharwal_guiding_2012}, where a Multi-Armed Bandit is used to model the tree search.

Feature gating has a long and successful history in machine learning (see \citeauthor{FeatureGatingSurvey}), ranging from LSTMs \cite{LSTM} to GRUs \cite{GRU}. 
The idea of using a tree state to drive a feature gating of the branching variables is an example of top-down modulation, which has been shown to perform well in other deep learning applications \cite{BeyondSkip, TopDownPyramid, ExGate}. 
With respect to learning across non-static action spaces, the most similar to our work is \citeauthor{ChangingActionSetLifeLong}, in the continual learning setting. 
Unlike the traditional Markov Decision Process formulation of RL, the input to our policies is not a generic state but rather includes a parameterized hand-crafted representation of the available actions, thus continual learning is not a relevant concern for our framework. 
Other works from the RL setting learn representations of \emph{static} action spaces \cite{LargeDiscreteRLActionSpaces, LearningTheActionSpace}, while in contrast the action space of BVS changes dynamically with $|\mathcal{C}_t|$. 

\section{Conclusions and Future Directions}
\label{sec:conclusions}

Branching variable selection is a crucial factor in B\&B success, and we setup a novel imitation learning framework to address it. 
We sought to learn branching policies that generalize across heterogeneous MILPs, regardless of the instances' structure and formulation size. 
In doing so, we undertook a step towards a broader type of generalization. 
The novelty of our approach is relevant for both the ML and the MILP worlds. 
On the one hand, we developed parameterizations of the candidate variables and of the search trees, and designed a DNN architecture that handles candidate sets of varying size. 
On the other hand, the data encoded in our $\mathit{Tree}_t$ parameterization is not currently exploited by state-of-the-art MILP solvers, but we showed that this type of information could indeed help in adapting the branching criteria to different search dynamics. 
Our results on MILP benchmark instances clearly demonstrated the advantage of incorporating a search-tree context to modulate BVS and aid generalization to heterogeneous problems, in terms of both better test accuracy and smaller explored B\&B trees. 
The comparison with the GCNN setup of \citeauthor{NIPS2019_9690} reinforced our conclusions: experiments showcased the inability of the GCNN paradigm alone to generalize to new instances for which no analogs were available during training. 
One crucial step towards improving over state-of-the-art solvers is precisely that of being able to generalize across heterogeneous problems, and our work is the first paper in the literature attaining this target.

There surely are additional improvements to be gained by continuing to experiment with IL methods for branching, and also by exploring innovative RL settings. 
Indeed, the idea and the benefits of using an explicit parameterization of B\&B search trees -- which we demonstrated in the IL setup -- could be expanded even more in the RL one, for both state representations and the design of branching rewards.

\section*{Acknowledgements}
We would like to thank Ambros Gleixner, Gerald Gamrath, Laurent Charlin, Didier Chételat, Maxime Gasse, Antoine Prouvost, Leo Henri and Sébastien Lachapelle for helpful discussions on the branching framework. We also thank Compute Canada for compute resources. This work was supported by CIFAR and IVADO. We also thank the anonymous reviewers for their helpful feedback.

\small
\bibliography{main}

\appendix

\section{Dataset curation}
\label{app:dataset}

To curate a dataset of heterogeneous MILP instances, we consider the standard benchmark libraries MIPLIB 3, 2010 and 2017 \cite{miplib3,MIPLIB2010,miplib2017}, together with the collection of \citeauthor{milplib}. We assess the problems by analyzing B\&B roll-outs of SCIP with its default branching rule (\texttt{relpscost}) and a \texttt{random} one, enforcing a time limit of 1h in the same solver setting used for our experiments (see Appendix~\ref{app:setting}). 
We focus on instances whose tree exploration is on average relatively contained (in the tens/hundreds of thousands nodes, maximum) and whose optimal value is known. This choice is primarily motivated by the need of ensuring a fair comparison among branching policies in terms of tree size, which is more easily achieved when roll-outs do not hit the time-limit. We also remove problems that are solved at the root node (i.e., those for which no branching was performed).

Final training and test sets comprise 19 and 8 instances, respectively, for a total of 27 problems. They are summarized in Table~\ref{tab:milp_instances}, where we report their size, the number of binary/integer/continuous variables, the number of constraints, their membership in the train/test split and their library of origin. The constraints of each problem are of different types and give rise to various structures.

\begin{table*}[t]
    \caption{The curated MILP dataset. For each instance we report: the number of variables (Vars) and their types (binary, integers and continuous), the number of constraints (Conss), the membership in the train/test split and the library of origin.}
    \label{tab:milp_instances}
    \centering
    \begin{footnotesize}
    \begin{tabular}{lrcrcr}
\toprule
           Name         & Vars  &  Types (bin - int - cont)  & Conss &  Set & Library  \\
\toprule
    air04               & 8904  & 8904 - 0 - 0      & 823   & train & MIPLIB 3 \\
    air05               & 7195  & 7195 - 0 - 0      & 426   & train & MIPLIB 3\\
    dcmulti             & 548   & 75  - 0  - 473    & 290   & train & MIPLIB 3\\
    eil33-2             & 4516  & 4516 - 0 - 0      & 32    & train & MIPLIB 2010 \\
    istanbul-no-cutoff  & 5282  & 30 - 0 - 5252     & 20346 & train & MIPLIB 2017 \\
    l152lav             & 1989  & 1989 - 0 - 0      & 97    & train & MIPLIB 3 \\
    lseu                & 89    & 89  - 0  - 0      & 28    & train & MIPLIB 3 \\
    misc03              & 160   & 159  - 0 - 1      & 96    & train & MIPLIB 3 \\
    neos20              & 1165  & 937  - 30 - 198  & 2446  & train & MILPLib \\
    neos21              & 614   & 613 - 0 - 1       & 1085  & train & MILPLib \\
    neos-476283         & 11915 & 5588 - 0 - 6327   & 10015 & train & MIPLIB 2010 \\
    neos648910          & 814   & 748  - 0  - 66    & 1491  & train & MILPLib \\
    pp08aCUTS           & 240   & 64  - 0  - 176    & 246   & train & MIPLIB 3 \\
    rmatr100-p10        & 7359  & 100 - 0 - 7259    & 7260  & train & MIPLIB 2010 \\
    rmatr100-p5         & 8784  & 100 - 0 - 8684    & 8685  & train & MIPLIB 2010 \\
    sp150x300d          & 600   & 300 - 0 - 300     & 450   & train & MIPLIB 2017 \\
    stein27             & 27    & 27  - 0  - 0      & 118   & train & MIPLIB 3 \\
    swath1              & 6805  & 2306 - 0 - 4499   & 884   & train & MIPLIB 2017 \\ 
    vpm2                & 378   & 168 - 0 - 210     & 234   & train & MIPLIB 3 \\
\midrule
    map18               & 164547& 146 - 0 - 164401  & 328818& test & MIPLIB 2010 \\
    mine-166-5          & 830   & 830 - 0 - 0       & 8429  & test & MIPLIB 2010 \\ 
    neos11              & 1220  & 900 - 0 - 320     & 2706  & test & MILPLib\\
    neos18              & 3312  & 3312 - 0 - 0      & 11402 & test & MIPLIB 2010 \\
    ns1830653           & 1629  & 1458 - 0 - 171    & 2932  & test & MIPLIB 2010 \\
    nu25-pr12           & 5868  & 5832 - 36 - 0     & 2313  & test & MIPLIB 2017 \\
    rail507             & 63019 & 63009 - 0 - 10   & 509   & test & MIPLIB 2010 \\
    seymour1            & 1372  & 451 - 0 - 921     & 4944  & test & MIPLIB 2017\\
\bottomrule
\end{tabular}
\end{footnotesize}
\end{table*}

\section{Hand-crafted input features}
\label{app:features}

Hand-crafted input features for candidate variables ($C_t$) and tree state ($\mathit{Tree}_t$) are reported in Table~\ref{tab:features}. To ease their reading, we present them subdivided in groups, and synthetically describe them by the SCIP API functions with which they are computed. We make use of different functions to normalize and compare the solver inputs.

To compute the branching scores $s_i$ of a candidate variable $i\in\mathcal{C}_t$, with respect to an average score $s_\mathit{avg}$, we use the formula implemented in SCIP \texttt{relpscost} \cite{relpscost}:
\begin{equation*}
    \texttt{varScore}(s_i, s_\mathit{avg}) = 1 - \left(\frac{1}{1 + s_i/\max\{s_\mathit{avg}, 0.1\}}\right).
\end{equation*}
As in \cite{Achterberg2007}, we normalize inputs that naturally span different ranges by the following:
\begin{equation*}
    \texttt{gNormMax}(x) = \max\left\{\frac{x}{x+1}, 0.1\right\}.
\end{equation*}
To compare commensurable quantities (e.g., upper and lower bounds), we compute measures of relative distance and relative position:
\begin{equation*}
    \texttt{relDist}(x, y) = \begin{cases} 
    0 & \text{, if } xy<0 \\
    \frac{|x-y|}{\max\{|x|, |y|, \num{1e-10}\}} & \text{, else}
    \end{cases}
\end{equation*}
\begin{equation*}
    \texttt{relPos}(z,x,y) = \frac{|x-z|}{|x-y|}.
\end{equation*}
We also make use of usual statistical functions such as \texttt{min}, \texttt{max}, \texttt{mean}, standard deviation \texttt{std} and 25-75\% quantile values (denoted in Table~\ref{tab:features} as \texttt{q1} and \texttt{q3}, respectively).

Further information on each feature can be gathered by searching the SCIP online documentation at \url{https://scip.zib.de/doc-6.0.1/html/}.

\section{Solver setting and hardware}
\label{app:setting}

Regarding the MILP solver parametric setting, we use SCIP 6.0.1 and set a time-limit of 1h on all B\&B evaluations. We leave on presolve routines and cuts separation (as in default mode), while disabling \emph{all} primal heuristics and reoptimization (also off at default). To control SB side-effects and properly compute the \emph{fair} number of nodes \cite{gamrath_impact}, we additionally turn off SB conflict analysis and the use of probing bounds identified during SB evaluations. We also disable feasibility checking of LP solutions found during SB with propagation, and always trigger the reevaluation of SB values. Finally, the known optimal solution value is provided as cutoff to each model, and a random seed determines variables' permutations. Parameters are summarized in Table \ref{tab:params}.

To benchmark the GCNN model of \citeauthor{NIPS2019_9690}, we do not use the original parametric setting of \citeauthor{NIPS2019_9690} but the one summarized above. For the rest, data collection and experiments are executed as in the original paper, with full SB used as expert rule. Note that a time-limit of 10 minutes is enforced in data-collection runs. To train and test \textsc{gcnn}, we had to reduce the batch size parameter from 32 to 4 in training, and from 128 to 16 in test, in order to avoid memory issues, as our MILP dataset contains bigger instances than those used in \citeauthor{NIPS2019_9690}. Finally, the state buffer was deactivated at evaluation time due to the presence of cuts in our solver setting. 

For the IL experiments, we used the following hardware: Two Intel Core(TM) i7-6850K CPU @ 3.60GHz, 16GB RAM and an NVIDIA TITAN Xp 12GB GPU. Evaluations of SCIP branching rules ran on dual Intel(R) Xeon(R) Gold 6142 CPU @ 2.60GHz, equipped with 512GB of RAM. The entire benchmark of \textsc{gcnn} was executed on Intel(R) Xeon(R) Gold 6126 CPU @ 2.60GHz and an Nvidia Tesla V100 GPU.

\begin{table}[h]
    \caption{SCIP parametric setting.}
    \label{tab:params}
    \centering
    \begin{footnotesize}
    \begin{tabular}{l}
    \toprule
    limits/time = 3600\\
    presolving/maxrounds = -1\\
    separating/maxrounds = -1\\
    separating/maxroundsroot = -1\\
    heuristics/*/freq = -1\\
    reoptimization/enable = False\\
    conflict/usesb = False\\
    branching/fullstrong/probingbounds = False\\
    branching/relpscost/probingbounds = False\\
    branching/checksol = False\\
    branching/fullstrong/reevalage = 0\\
    model.setObjlimit(cutoff\_value)\\
    randomization/permutevars = True\\
    randomization/permutationseed = scip\_seed\\
    \bottomrule
    \end{tabular}
    \end{footnotesize}
\end{table}

\section{Data augmentation}
\label{app:data-aug}

To augment our dataset, we (i) run MILP instances with different random seeds to exploit performance variability \cite{LodiPerfVar}, and (ii) perform $k$ random branchings at the top of the tree, before switching to the default SCIP branching rule and collect data. To quantify the effects of such operations in diversifying the search trees, we compute coefficients of variations of performance measurements \cite{MIPLIB2010}. In particular, assuming performance measurements $n_l, l=1\dots L$ are available, we compute the variability score $\mathit{VS}$ as
\begin{equation}
    \mathit{VS} = \frac{L}{\sum_{l=1}^L n_l}\sqrt{\sum_{l=1}^L \left(n_l - \frac{\sum_{l=1}^L n_l}{L}\right)^2}.
\end{equation}
Table~\ref{tab:k_vs} reports such coefficients for all instances, using as performance measures the number of nodes explored in the five runs from (i). The observed coefficients range in $[0.03, 1.70]$: the majority of the instances presents a variability of at least 0.20, confirming (i) as an effective way of diversifying our dataset. Similarly, we report the shifted geometric means of the number of nodes over the five runs for each $k\in\{0, 1, 5, 10, 15\}$, and additionally compute the variability of those means, across different $k$'s ($\mathit{VS}_k$). Generally, the size of the explored trees grows with $k$, i.e., initial random branchings affect the nodes' count for worse -- though the opposite can also happen in few cases. The coefficients of variation of the nodes shifted geometric means across different $k$'s range in $[0.07, 0.79]$ in the training set, so (ii) also appears effective for data augmentation. Overall, both (i) and (ii) appear effective ways of diversifying our dataset.

\begin{table*}[t]
    \caption{Variability scores $\mathit{VS}$ are reported for \texttt{relpscost}, and are computed using the 5 runs with $k=0$ (i.e., SCIP default runs, over seeds $\{0, \dots, 4\}$). Total number of nodes explored by data collection runs with $k$ random branchings, in shifted geometric means over 5 runs is also reported. Finally, $\mathit{VS}_k$ is the coefficient of variation of the five means, across different $k$'s. }
    \label{tab:k_vs}
    \centering
    \begin{footnotesize}
\begin{tabular}{ll|r|rrrrr|r}
\toprule
           Instance &    Set &    $\mathit{VS}$ &    $k=0$ &    $k=1$ &    $k=5$ &   $k=10$ &   $k=15$ &  $\mathit{VS}_k$ \\
\midrule
              air04 &  train &             0.20 &     8.19 &    12.02 &    46.57 &    85.35 &   119.49 &             0.79 \\
              air05 &  train &             0.26 &    60.25 &    61.07 &   115.94 &   196.27 &   274.44 &             0.59 \\
            dcmulti &  train &             0.21 &     9.38 &    13.75 &    27.53 &    34.99 &    45.00 &             0.50 \\
            eil33-2 &  train &             0.69 &   583.34 &   648.47 &   492.95 &   531.37 &   441.24 &             0.13 \\
 istanbul-no-cutoff &  train &             0.11 &   242.39 &   234.01 &   271.35 &   279.38 &   311.39 &             0.10 \\
            l152lav &  train &             0.36 &    10.14 &    16.54 &    29.31 &    55.51 &    61.14 &             0.59 \\
               lseu &  train &             0.43 &   148.99 &   152.65 &   154.16 &   182.55 &   177.35 &             0.09 \\
             misc03 &  train &             0.38 &    12.11 &    10.59 &    13.80 &    22.59 &    31.80 &             0.44 \\
             neos20 &  train &             1.22 &   200.26 &   282.68 &   557.15 &   434.03 &   944.75 &             0.54 \\
             neos21 &  train &             0.15 &   668.44 &   771.77 &   898.79 &  1110.82 &  1158.07 &             0.21 \\
         neos648910 &  train &             0.60 &    39.83 &    48.16 &    65.84 &    41.05 &    59.72 &             0.20 \\
        neos-476283 &  train &             0.48 &   204.88 &   219.58 &   384.86 &   480.37 &   715.78 &             0.47 \\
          pp08aCUTS &  train &             0.31 &    69.66 &    80.39 &    92.60 &    69.94 &    76.43 &             0.11 \\
        rmatr100-p5 &  train &             0.04 &   411.93 &   419.21 &   451.01 &   461.83 &   494.09 &             0.07 \\
       rmatr100-p10 &  train &             0.03 &   806.35 &   799.24 &   860.60 &   933.80 &   965.07 &             0.08 \\
         sp150x300d &  train &             1.70 &   182.22 &   462.45 &   484.55 &   483.89 &   439.69 &             0.28 \\
            stein27 &  train &             0.42 &   926.82 &  1062.69 &  1098.41 &  1162.57 &  1154.01 &             0.08 \\
             swath1 &  train &             0.53 &   298.58 &   280.49 &   230.12 &   256.84 &   267.55 &             0.09 \\
               vpm2 &  train &             0.19 &   199.46 &   180.93 &   275.57 &   273.33 &   316.82 &             0.20 \\
\midrule
              map18 &   test &             0.09 &   270.25 &   309.77 &   401.79 &   447.34 &   489.85 &             0.21 \\
         mine-166-5 &   test &             0.82 &   175.10 &    70.77 &   642.33 &   942.63 &  1619.75 &             0.81 \\
             neos11 &   test &             0.30 &  2618.27 &  3114.62 &  3488.40 &  2898.41 &  2659.96 &             0.11 \\
             neos18 &   test &             0.53 &  2439.29 &  2747.77 &  4061.40 &  4655.59 &  5714.05 &             0.31 \\
          ns1830653 &   test &             0.09 &  3489.07 &  3913.58 &  4091.59 &  4839.39 &  4772.73 &             0.12 \\
          nu25-pr12 &   test &             1.18 &    21.39 &    16.97 &    56.04 &   101.34 &   119.05 &             0.66 \\
            rail507 &   test &             0.08 &   543.39 &   562.09 &   854.76 &  1207.15 &  1196.33 &             0.33 \\
           seymour1 &   test &             0.07 &   866.32 &  1174.18 &  1825.04 &  2739.45 &  3313.87 &             0.47 \\
\bottomrule
\end{tabular}
\end{footnotesize}
\end{table*}

\section{IL optimization dynamics}
\label{app:il-opt}

\begin{figure*}[t]
    \centering
    \subfigure[\label{fig:train-loss}]{\includegraphics[height=4.5cm]{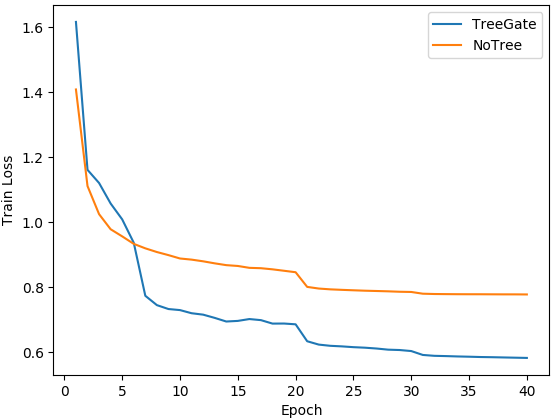}} \quad
    \subfigure[\label{fig:val-loss}]{\includegraphics[height=4.5cm]{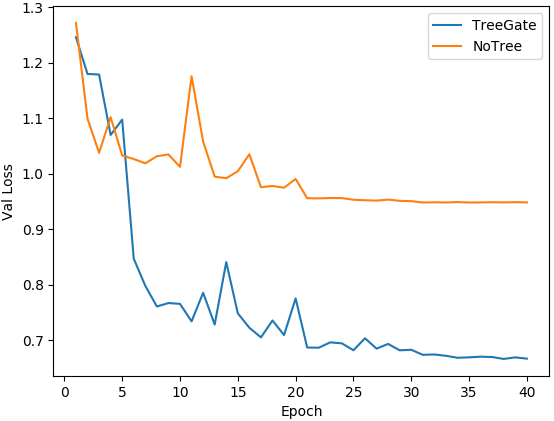}}
    \caption{(a) Train and (b) validation loss curves for the best NoTree (orange) and TreeGate (blue) policies.}
    \label{fig:loss-plots}
\end{figure*}

\begin{figure*}[t]
    \centering
    \subfigure[\label{fig:val-1}]{\includegraphics[height=4.5cm]{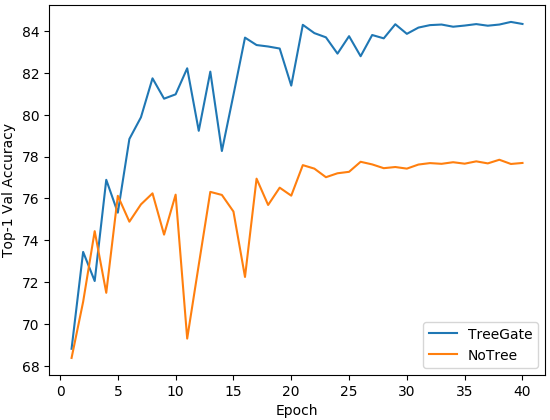}} \quad
    \subfigure[\label{fig:val-5}]{\includegraphics[height=4.5cm]{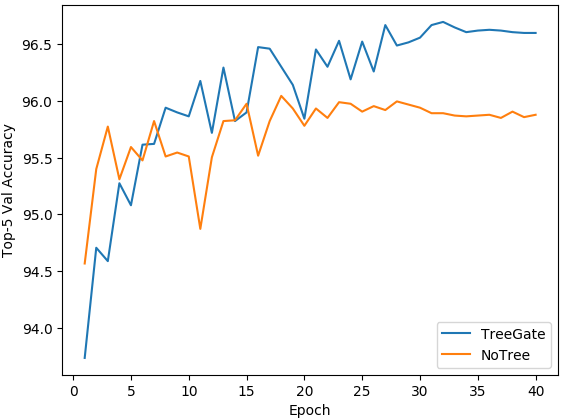}}
    \caption{(a) Validation top-1 and (b) top-5 accuracy plots for the best NoTree (orange) and TreeGate (blue) policies.}
    \label{fig:acc-plots}
\end{figure*}

\paragraph{Best policies}{We report hyper-parameters and performance details of the best learned NoTree and TreeGate policies in Table~\ref{tab:best_policies}. The top-1 test accuracy averages at $65.90 \pm 1.6$ for the NoTree models, while TreeGate ones score at $83.08 \pm 0.86$; the gap in validation accuracy is also significant. In terms of B\&B roll-outs, NoTree models explore on average $1336.12 \pm 73.32$ nodes, against the $1126.97 \pm 46.85$ of TreeGate ones. 
What we observe is that best test accuracy does not necessarily translate into best solver performance. The NoTree policy with the best solver performance exhibits an approximately 4\% gap from the optimal top-1 test accuracy model, but an improvement over 7\% in solver performance. In the case of TreeGate, the best model corresponds to that realizing the best top-1 test accuracy (83.70\%), and brings a 19\% (resp. 7\%) improvement over the NoTree policy, in top-1 (resp. top-5) test accuracy.
Additionally, we present plots of the optimization dynamics for the selected NoTree and TreeGate policies. Figure~\ref{fig:loss-plots} shows the training loss curves, while Figure~\ref{fig:acc-plots} depicts top-1 and top-5 validation accuracy curves. In general, we see that the TreeGate policy enjoys a better conditioned optimization. Note however that for top-5 validation accuracy the two policies are quite close.
}

\begin{table*}[t]
    \caption{Best trained NoTree and TreeGate models. For each policy, we report the corresponding hyper-parameters, top-1 and top-5 test and validation accuracy scores, and shifted geometric means of B\&B nodes over \textsc{All}, \textsc{Train} and \textsc{Test} instances. 
    Policies selected as best ones are boldfaced.}
    \label{tab:best_policies}
    \centering
    \begin{footnotesize}
    \begin{tabular}{llccccc}
\toprule
          Policy & $h$ / $d$ / $\mathit{LR}$ & Test acc@1 (@5) & Val acc@1 (@5) &  \textsc{All} & \textsc{Train} & \textsc{Test}  \\
\toprule
    \multirow{5}{*}{NoTree} & 32 / -- / 0.0001 & 68.37 (91.43) & 75.40 (95.23) & 1341.72 & 859.17 & 3695.04 \\
     & 64 / -- / 0.0001 & 67.05 (89.18) & 76.45 (95.11) &  1363.73 & 847.63 & 4010.65 \\
     & 128 / -- / 0.0001 & 65.44 (90.21) & 76.77 (95.66) &  1454.20 & 875.19 & 4601.72 \\
     & \textbf{128} / -- / \textbf{0.001} & \textbf{64.02 (88.51)} & \textbf{77.69 (95.88)} & \textbf{1241.79} & \textbf{834.40} & \textbf{3068.96} \\
     & 256 / -- / 0.0001 & 64.59 (90.13) & 77.29 (96.08) & 1279.18 & 731.16 & 4491.64 \\
    \midrule
    \multirow{4}{*}{TreeGate} & \textbf{64} / \textbf{5} / \textbf{0.01} & \textbf{83.70 (95.83)} & \textbf{84.33 (96.60)} & \textbf{1056.79} & \textbf{759.94} & \textbf{2239.47} \\
     & 256 / 2 / 0.001 & 83.69 (95.18) & 84.10 (96.42) & 1135.28 & 822.80  & 2369.35 \\
     & 32 / 3 / 0.01 & 83.31 (95.72) & 84.02 (96.50) & 1188.48 & 809.18  & 2849.28 \\
     & 128 / 5 / 0.001 & 81.61 (95.81) & 84.96 (96.74) & 1127.31 & 771.60 & 2666.73 \\
\bottomrule
\end{tabular}
\end{footnotesize}
\end{table*}

\paragraph{\textsc{gcnn}}{Detailed training, validation and test metrics for \textsc{gcnn} are reported in Table~\ref{tab:gcnn_details}.
}

\begin{table*}[t]
    \caption{Training and test details of the five \textsc{gcnn} models. All training phases finished by early stopping, after 20 epochs without improvement.}
    \label{tab:gcnn_details}
    \centering
    \begin{footnotesize}
    \begin{tabular}{ccccccc}
\toprule
Seed & \# epochs & Best valid. loss & acc@1 & acc@5 & Test acc@1 & Test acc@5\\
\midrule
0 &	142	&	4.472 &	0.192 &	0.383 &	0.1327 & 0.3828\\
1 &	206	&	4.455 &	0.197 &	0.393 &	0.1577 &	0.4738\\
2 &	143	&	4.472 &	0.186 &	0.378 &	0.1575 &	0.5068\\
3 &	211	&	4.465 &	0.189 &	0.383 &	0.2206 &	0.5004\\
4 &	269	&	4.455 &	0.200	&	0.385 &	0.0954 &	0.3443\\
\bottomrule
\end{tabular}
\end{footnotesize}
\end{table*}

\begin{figure*}[t]
    \centering
    \subfigure[\label{fig:bn-train}]{\includegraphics[height=4.5cm]{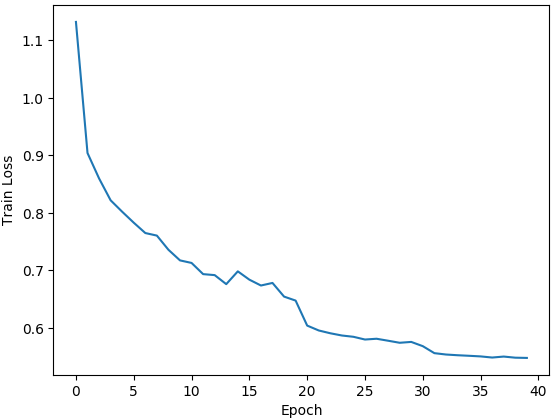}} \quad
    \subfigure[\label{fig:bn-val}]{\includegraphics[height=4.5cm]{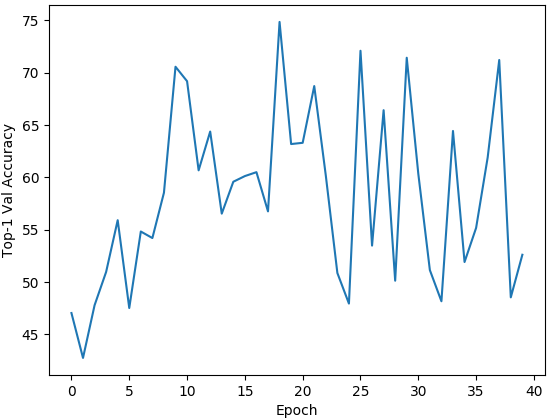}}
    \caption{(a) Train loss and (b) validation top-1 Accuracy for NoTree+BN policy with hidden size $h=64$, $\mathit{LR}=0.001$.}
    \label{fig:notree-bn-plots}
\end{figure*}

\paragraph{Instability of batch-norm}{As observed in Figures~\ref{fig:loss-plots} and ~\ref{fig:acc-plots}, optimization dynamics for NoTree seem to be of a much slower nature than those of TreeGate. One common option to speed up training is to use batch normalization (BN) \cite{BatchNorm}. In our architectures for branching, one may view the cardinality of the candidate sets $|\mathcal{C}_t|$ as a batch dimension. When learning to branch across heterogeneous MILPs, such batch dimension can (and will) vary by orders of magnitude. Practically, our dataset has $|\mathcal{C}_t|$ varying from $<10$ candidates to over 300. To this end, BN has been shown to struggle in the small-batch setting \cite{GN}, and in general we were unsure of the reliability of BN with such variable batch-sizes. 

Indeed, in our initial trials with BN we observed highly unreliable performance. Two troubling outcomes emerge when using BN in our NoTree policies: 1) the validation accuracy varies wildly, as shown in Figure~\ref{fig:notree-bn-plots}, or 2) the NoTree+BN policy exhibits a stable validation accuracy curve, but would time-limit on \emph{train instances}, i.e., would perform poorly in terms of solver performance. In particular, case 2) happened for a NoTree+BN policy with hidden size $h=64$ and $\mathit{LR}=0.001$, reaching the 1h time-limit on train instance neos-476283, over all five runs (on different seeds); the geometric mean of explored nodes was $66170.66$. We remark that in our non-BN experiments, all of our trained policies (both TreeGate and NoTree) managed to solve all the train instances without even coming close to time-limiting. Moreover, none of our training and validation curves ever remotely resemble those in Figure~\ref{fig:bn-val}.

For these reasons we opted for a more streamlined presentation of our results, without BN in the current framework. We leave it for future work to analyze the relationship between the nature of local minima in the IL optimization landscape and solver performance.
}

\paragraph{SCIP evaluation}{
Instance-specific details of SCIP evaluations of all policies are reported in Table~\ref{tab:nnodes_eval}. Results are less clear-cut from an instance-wise perspective, with neither policy emerging as an absolute winner. Nonetheless, TreeGate is at least 10\% (resp. 25\%) better than NoTree on 10 (resp. 8) instances, while the opposite only happens 6 (resp. 3) times. In this sense, the reductions in tree sizes achieved by TreeGate are overall more pronounced.
}

\begin{table*}[!t]
    \caption{Total number of nodes explored by learned and SCIP policies for train and test instances, in shifted geometric means over 5 runs on seeds $\{0, \dots, 4\}$. We mark with * the cases in which time-limits were hit. For \texttt{relpscost}, we also compute the \emph{fair} number of nodes. Aggregated measures are over the entire dataset (\textsc{All}), as well as over \textsc{Train} and \textsc{Test} sets.}
    \label{tab:nnodes_eval}
    \centering
    \begin{footnotesize}
\begin{tabular}{lrrrrrr}
\toprule
           Instance &    NoTree &  TreeGate &  \textsc{gcnn} &     \texttt{random} &     \texttt{pscost} &  \texttt{relpscost} (fair) \\
\midrule
       \textsc{All} &    1241.79 &   1056.79 &  *3660.32 &    *6580.79 &    *1471.61 &     286.15 (719.20) \\
     \textsc{Train} &     834.40 &    759.94 &   *1391.41 &    *2516.04 &     884.37 &     182.27 (558.34) \\
      \textsc{Test} &    3068.96 &   2239.47 &  *33713.63 &   *61828.29 &    *4674.34 &     712.77 (1276.76) \\
\midrule
              air04 &      645.99 &    536.07 & 1249.11  &    6677.96 &     777.65 &       8.19 (114.39) \\
              air05 &      789.70 &    516.06 &  3318.40 &   12685.83 &    1158.89 &      60.25 (277.22) \\
            dcmulti &      203.53 &    187.49 &  160.02 &     599.12 &     122.39 &       9.38 (68.30) \\
            eil33-2 &     7780.85 &   8767.27 &  24458.14 &   12502.02 &    8337.63 &     583.34 (9668.71) \\
 istanbul-no-cutoff &      447.26 &    543.71 &  446.93 &    1085.16 &     613.68 &     242.39 (328.25) \\
            l152lav &      621.82 &    687.91 &  1649.27 &    6800.06 &     964.53 &      10.14 (250.04) \\
              lseu &      372.67 &    396.71 &  316.37 &     396.73 &     375.31 &     148.99 (389.88) \\
             misc03 &      241.40 &    158.39 &  957.10 &     118.37 &     151.07 &      12.11 (294.11) \\
             neos20 &     2062.23 &   1962.95 &  507.95 &   10049.15 &    2730.01 &     200.26 (612.75) \\
             neos21 &     1401.84 &   1319.73 &  770.97 &    7016.55 &    1501.54 &     668.44 (1455.29) \\
         neos648910 &      140.05 &    175.82 &  162.24 &    1763.05 &    1519.01 &      39.83 (166.53) \\
        neos-476283 &    13759.59 &   6356.81 & *21077.30  &   *94411.77 &    2072.84 &     204.88 (744.65) \\
          pp08aCUTS &      267.86 &    293.74 &  327.33 &     337.76 &     271.92 &      69.66 (350.21) \\
        rmatr100-p5 &      443.35 &    460.48 &  747.91 &    1802.38 &     451.71 &     411.93 (785.15) \\
      rmatr100-p10 &      908.27 &    906.04 &  1169.40 &    4950.77 &     894.65 &     806.35 (1214.76) \\
         sp150x300d &      868.60 &    785.27 &  50004.09 &    1413.64 &     991.52 &     182.22 (300.42) \\
            stein27 &     1371.44 &   1146.79 &  3093.62 &    1378.91 &    1322.36 &     926.82 (1111.25) \\
             swath1 &     1173.14 &   1165.39 &  1690.07 &    1429.21 &    1107.52 &     298.58 (2485.63) \\
              vpm2 &      589.03 &    440.74 &  313.71 &     594.62 &     546.45 &     199.46 (463.12) \\
\midrule
              map18 &       457.89 &    575.92 &  *3907.64 &   11655.33 &    1025.74 &     270.25 (441.18) \\
         mine-166-5 &      3438.44 &   4996.48 &  *233142.25 &  *389437.62 &    4190.41 &     175.10 (600.22) \\
             neos11 &      3326.32 &   3223.46 &  1642.07 &   29949.69 &    4728.49 &    2618.27 (5468.05) \\
             neos18 &     15611.63 &  10373.80 &  40794.74 &  228715.62 &  *133437.40 &    2439.29 (5774.36) \\
          ns1830653 &      6422.37 &   5812.03 &  *22931.45 &  288489.30 &   12307.90 &    3489.07 (4311.84) \\
          nu25-pr12 &       357.00 &     86.80 &  *45982.34 &    1658.41 &     342.47 &      21.39 (105.61) \\
            rail507 &      9623.05 &   3779.05 &  *75663.48 &   *80575.84 &    4259.98 &     543.39 (859.37) \\
           seymour1 &      3202.20 &   1646.82 &  *319046.04 &  *167725.65 &    3521.47 &     866.32 (1096.67) \\
\bottomrule
\end{tabular}
\end{footnotesize}
\end{table*}

\onecolumn
\begin{center}
\begin{footnotesize}
\LTcapwidth=\textwidth
\begin{longtable}{ll}
\caption{Description of the hand-crafted input features. Features are doubled [x2] when they are computed for both upward and downward branching directions. For features about open nodes, open\_lbs denotes the list of lower bound estimates of the open nodes, while open\_ds the list of depths across open nodes. Table continues on the next page.}
  \label{tab:features} \\
\toprule
\textbf{Group description (\#)} & \textbf{Feature formula} (SCIP API) \\
\midrule
\endfirsthead

\multicolumn{2}{c}%
{{\tablename\ \thetable{} -- continued from previous page}} \\
  \toprule 
  \textbf{Group description (\#)} & \textbf{Feature formula} (SCIP API) \\
  \midrule
\endhead

\hline \hline
\endlastfoot

\emph{Candidate state $[C_t]_i, i\in\mathcal{C}_t$} & \\
\midrule
General solution (2) & SCIPvarGetLPSol \\
& SCIPvarGetAvgSol \\
Branchings depth (2) & 1 - (SCIPvarGetAvgBranchdepthCurrentRun / SCIPgetMaxDepth) [x2]\\
Branching scores (5) & \texttt{varScore}(SCIPgetVarConflictScore, SCIPgetAvgConflictScore) \\
& \texttt{varScore}(SCIPgetVarConflictlengthScore, SCIPgetAvgConflictlengthScore)\\
& \texttt{varScore}(SCIPgetVarAvgInferenceScore, SCIPgetAvgInferenceScore) \\
& \texttt{varScore}(SCIPgetVarAvgCutoffScore, SCIPgetAvgCutoffScore)\\
& \texttt{varScore}(SCIPgetVarPseudocostScore, SCIPgetAvgPseudocostScore)\\
PC stats (6) & SCIPgetVarPseudocostCountCurrentRun / SCIPgetPseudocostCount [x2]\\
& SCIPgetVarPseudocostCountCurrentRun / SCIPvarGetNBranchingsCurrentRun [x2]\\
& SCIPgetVarPseudocostCountCurrentRun / branch\_count [x2]\\
Implications (2) & SCIPvarGetNImpls [x2] \\
Cliques (2) & SCIPvarGetNCliques / SCIPgetNCliques [x2]\\
Cutoffs (2) & \texttt{gNormMax}(SCIPgetVarAvgCutoffsCurrentRun) [x2]\\
Conflict length (2) & \texttt{gNormMax}(SCIPgetVarAvgConflictlengthCurrentRun) [x2]\\
Inferences (2) & \texttt{gNormMax}(SCIPgetVarAvgInferencesCurrentRun) [x2]\\
\midrule
\emph{Search tree state $\mathit{Tree}_t$} & \\
\midrule
Current node (8) & SCIPnodeGetDepth / SCIPgetMaxDepth\\
& SCIPgetPlungeDepth / SCIPnodeGetDepth\\
& \texttt{relDist}(SCIPgetLowerbound, SCIPgetLPObjval)\\
& \texttt{relDist}(SCIPgetLowerboundRoot, SCIPgetLPObjval)\\
& \texttt{relDist}(SCIPgetUpperbound, SCIPgetLPObjval)\\
& \texttt{relPos}(SCIPgetLPObjval, SCIPgetUpperbound, SCIPgetLowerbound)\\
& \texttt{len}(getLPBranchCands) / getNDiscreteVars\\
& nboundchgs / SCIPgetNVars\\
Nodes and leaves (8) & SCIPgetNObjlimLeaves / nleaves\\
& SCIPgetNInfeasibleLeaves / nleaves\\
& SCIPgetNFeasibleLeaves / nleaves\\
& (SCIPgetNInfeasibleLeaves + 1) / (SCIPgetNObjlimLeaves + 1)\\
& SCIPgetNNodesLeft / SCIPgetNNodes \\
& nleaves / SCIPgetNNodes\\
& ninternalnodes / SCIPgetNNodes \\
& SCIPgetNNodes / ncreatednodes \\
Depth and backtracks (4) & nactivatednodes / SCIPgetNNodes\\
& ndeactivatednodes / SCIPgetNNodes \\
& SCIPgetPlungeDepth / SCIPgetMaxDepth \\
& SCIPgetNBacktracks / SCIPgetNNodes \\
LP iterations (4) & \texttt{log}(SCIPgetNLPIterations / SCIPgetNNodes)\\
& \texttt{log}(SCIPgetNLPs / SCIPgetNNodes)\\
& SCIPgetNNodes / SCIPgetNLPs \\
& SCIPgetNNodeLPs / SCIPgetNLPs \\
Gap (4) & \texttt{log}(primaldualintegral)\\
& SCIPgetGap / lastsolgap \\
& SCIPgetGap / firstsolgap \\
& lastsolgap / firstsolgap \\
Bounds and solutions (5) & \texttt{relDist}(SCIPgetLowerboundRoot, SCIPgetLowerbound)\\
& \texttt{relDist}(SCIPgetLowerboundRoot, SCIPgetAvgLowerbound) \\
& \texttt{relDist}(SCIPgetUpperbound, SCIPgetLowerbound)\\
& SCIPisPrimalboundSol \\
& nnodesbeforefirst / SCIPgetNNodes \\
Average scores (12) & \texttt{gNormMax}(SCIPgetAvgConflictScore) \\
& \texttt{gNormMax}(SCIPgetAvgConflictlengthScore) \\
& \texttt{gNormMax}(SCIPgetAvgInferenceScore) \\
& \texttt{gNormMax}(SCIPgetAvgCutoffScore) \\
& \texttt{gNormMax}(SCIPgetAvgPseudocostScore) \\
& \texttt{gNormMax}(SCIPgetAvgCutoffs) [x2]\\
& \texttt{gNormMax}(SCIPgetAvgInferences) [x2] \\
& \texttt{gNormMax}(SCIPgetPseudocostVariance) [x2] \\
& \texttt{gNormMax}(SCIPgetNConflictConssApplied)\\
Open nodes bounds (12) & \texttt{len}(open\_lbs at \{\texttt{min}, \texttt{max}\}) / nopen [x2]\\
& \texttt{relDist}(SCIPgetLowerbound, \texttt{max}(open\_lbs))\\
& \texttt{relDist}(\texttt{min}(open\_lbs), \texttt{max}(open\_lbs))\\
& \texttt{relDist}(\texttt{min}(open\_lbs), SCIPgetUpperbound) \\
& \texttt{relDist}(\texttt{max}(open\_lbs), SCIPgetUpperbound) \\
& \texttt{relPos}(\texttt{mean}(open\_lbs), SCIPgetUpperbound, SCIPgetLowerbound) \\
& \texttt{relPos}(\texttt{min}(open\_lbs), SCIPgetUpperbound, SCIPgetLowerbound) \\
& \texttt{relPos}(\texttt{max}(open\_lbs), SCIPgetUpperbound, SCIPgetLowerbound) \\
& \texttt{relDist}(\texttt{q1}(open\_lbs), \texttt{q3}(open\_lbs))\\
& \texttt{std}(open\_lbs) / \texttt{mean}(open\_lbs) \\
& (\texttt{q3}(open\_lbs) - \texttt{q1}(open\_lbs)) / (\texttt{q3}(open\_lbs) + \texttt{q1}(open\_lbs)) \\
Open nodes depths (4) & \texttt{mean}(open\_ds) / SCIPgetMaxDepth \\
& \texttt{relDist}(\texttt{q1}(open\_ds), \texttt{q3}(open\_ds))\\
& \texttt{std}(open\_ds) / \texttt{mean}(open\_ds) \\
& (\texttt{q3}(open\_ds) - \texttt{q1}(open\_ds)) / (\texttt{q3}(open\_ds) + \texttt{q1}(open\_ds))\\
\end{longtable}
\end{footnotesize}
\end{center}
\twocolumn

\end{document}